\documentclass[letterpaper, 10 pt, conference]{ieeeconf}  

\IEEEoverridecommandlockouts                              

\usepackage{graphicx} 
\usepackage{mathtools}

\usepackage{amsmath} 

\usepackage{textcomp}
\usepackage{tabularx,booktabs}
\usepackage{color}
\usepackage[pdftex,dvipsnames]{xcolor}
\usepackage{array}
\usepackage{
algorithm} 
\usepackage{algpseudocode} 
\usepackage{subcaption} 

\newcolumntype{P}[1]{>{\centering\arraybackslash}p{#1}}
\newcolumntype{C}{>{\centering\arraybackslash}X} 

\usepackage{tikz}

\title{\LARGE \bf
Aggregating Multiple Bio-Inspired Image Region Classifiers For Effective And Lightweight Visual Place Recognition
}

\author{Bruno Arcanjo$^{1}$, Bruno Ferrarini$^{1}$, Maria Fasli$^{1}$, Michael Milford$^{2}$, Klaus D. McDonald-Maier$^{1}$ and Shoaib Ehsan$^{1, 3}$
\thanks{}
\thanks{$^{1}$B. Arcanjo, B. Ferrarini, K. D. McDonald-Maier and S. Ehsan are with the School of Computer Science and Electronic Engineering, University of Essex, United Kingdom {\tt\small (email: bq17319@essex.ac.uk; bferra@essex.ac.uk; mfasli@essex.ac.uk; kdm@essex.ac.uk; sehsan@essex.ac.uk)}}%
\thanks{$^{2}$M. Milford is with the School of Electrical Engineering and Computer Science, Queensland University of Technology, Brisbane, QLD 4000, Australia
        {\tt\small (email: michael.milford@qut.edu.au)}}%
\thanks{$^{3}$S. Ehsan is also with the school of Electronics and Computer Science, University of Southampton, United Kingdom \tt\small{(email: s.ehsan@soton.ac.uk)}}%
}

\begin{document}

\maketitle
\thispagestyle{empty}
\pagestyle{empty}

\begin{abstract}

Visual place recognition (VPR) enables autonomous systems to localize themselves 
within an environment using image information. While VPR techniques built upon a Convolutional Neural Network (CNN) 
backbone dominate state-of-the-art VPR performance, their high computational requirements make them unsuitable 
for platforms equipped with low-end hardware. Recently, a lightweight VPR system based on multiple bio-inspired classifiers, dubbed DrosoNets, has been proposed, achieving great computational efficiency at the cost of reduced absolute place retrieval performance. 
In this work, we propose a novel multi-DrosoNet localization system, dubbed RegionDrosoNet, with significantly improved VPR performance, while preserving a low-computational profile. 
Our approach relies on specializing distinct groups of 
DrosoNets on differently sliced partitions of the original images, increasing extrinsic model differentiation. Furthermore, we introduce 
a novel voting module to combine the outputs of all DrosoNets into the final place prediction which considers multiple top reference candidates from each DrosoNet. RegionDrosoNet outperforms other lightweight VPR techniques when dealing with both appearance changes and viewpoint variations. Moreover, it competes with computationally expensive methods on some benchmark datasets at a small fraction of their online inference time.

\end{abstract}

\section{Introduction}
\label{intro}

Visual place recognition (VPR) is an essential component of mobile robotics, as it allows the system to localize itself
in the runtime environment using only image data \cite{ref:vpr-survey}. The affordability and variety of camera sensors
makes VPR localization particularly attractive for hardware restricted robotic platforms, which are common in mobile
robotics \cite{ref:res_hardware_1}. Nevertheless, VPR is a complicated task and proposed solutions 
must deal with several visual challenges. The same place can appear vastly different when visited under different
illumination \cite{ref:illu_changes}, seasonal weather conditions \cite{ref:season_changes}, viewpoints \cite{ref:pov_changes}
and dynamic elements entering and leaving the scene \cite{ref:dyna_changes}. As alluded, mobile robotic platforms often operate
under low-end hardware, often due to physical size or monetary budget, making computational cost an added important consideration when designing VPR techniques \cite{ref:res_hardware_2}.
\begin{figure}[!t]
    \centering
    \includegraphics[width=0.9\columnwidth]{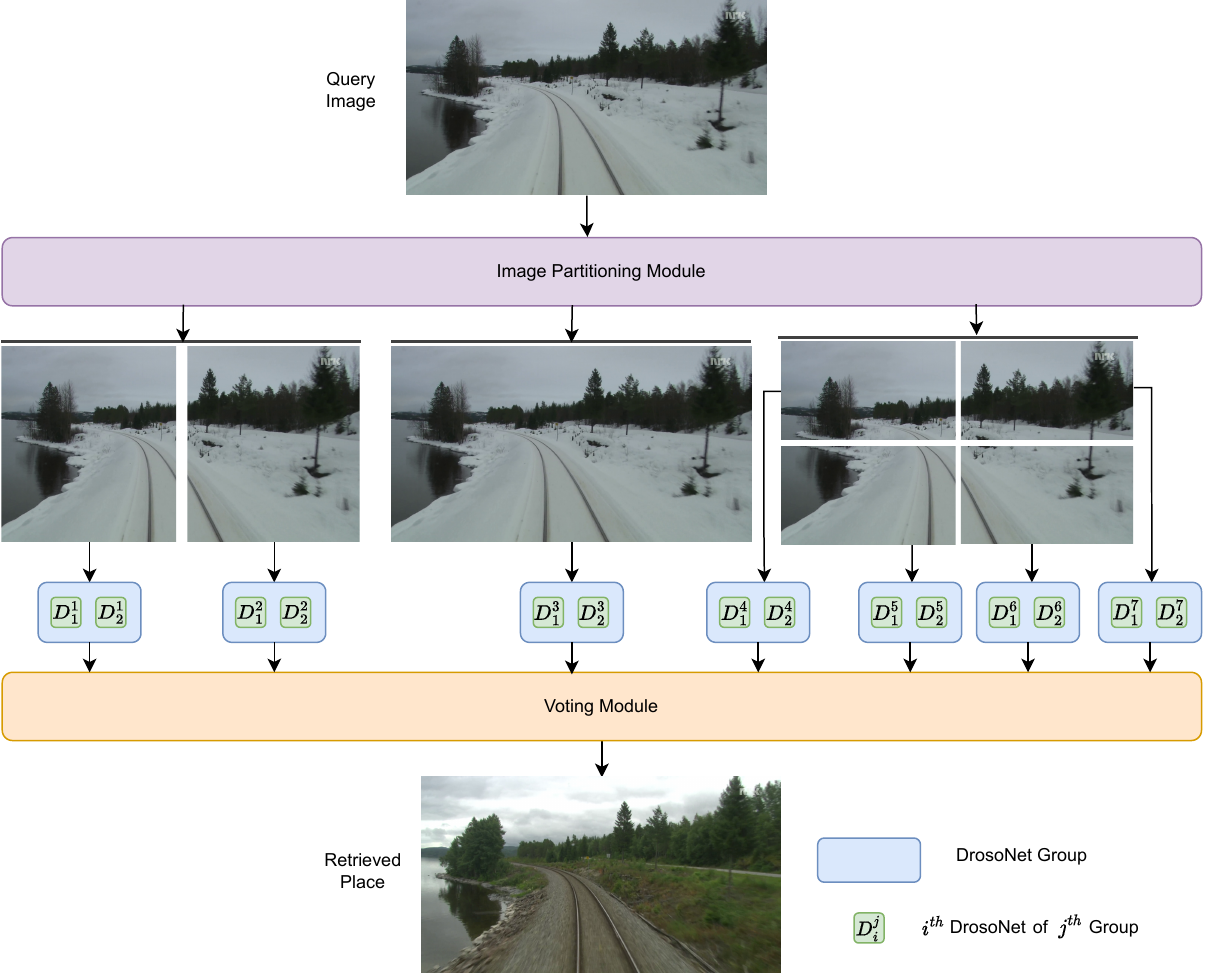}
    \caption{The query image is divided into multiple heterogeneous regions. Each region is then fed as input into a specialized DrosoNet group which was trained only
    on that particular region from the training set images. Finally, the output of each group is aggregated in 
    the voting module and a reference place is retrieved.}
    \label{fig:VA}
    \end{figure}
VPR methods based on Convolutional Neural Networks (CNNs) architectures have become increasingly popular due to 
their impressive performance. Indeed, visual features extracted from CNN layers achieve strong resilience against several
of the visual challenges intrinsic to VPR \cite{ref:cnn_for_vpr1}. However, as these networks grow deeper and more complex to 
achieve higher quality VPR, they also become less suitable for robotic setups equipped with heavily constricted hardware. 
Moreover, even if the hardware is able to support the use of an expensive CNN model in realtime, a lower computational 
demand is still valuable in saving power, allowing a mobile platform to operate for longer.

Recently, the authors proposed a lightweight VPR system \cite{voting_arcanjo} based on multiple bio-inspired voting units. 
Each unit, dubbed DrosoNet, is a compact neural network model inspired by the odour processing abilities of Drosophila 
Melanogaster (the common fruit fly) \cite{ref:droso_vpr}. The approach relies on the intrinsic parameter randomness 
that DrosoNet possesses, allowing for moderate unit differentiation and its extremely low
computational profile, allowing for a multi-DrosoNet system which is brought together with a voting mechanism attuned to VPR. Despite strong VPR performance relative to its computational efficiency, the absolute VPR quality of the system makes it unreliable in many of the tested environments, particularly when dealing with strong viewpoint variations.

In this work, we propose a novel multi-DrosoNet localization pipeline which achieves increased VPR performance across various visual challenges, while maintaining a low computational profile. The core of the approach, dubbed RegionDrosoNet, relies on introducing extrinsic model differentiation by training specialized DrosoNet groups on different regions of the training images. At inference time, as can be observed in Fig. \ref{fig:VA}, each partition of the query image is served as input to its respective group and each DrosoNet produces its reference place confidences. The training and inference process is tailored to DrosoNet, taking full advantage of its peculiarities: it's extremely fast and compact, allowing for the use of multiple units; it's a neural network classifier, not requiring a separate reference descriptor map for each image partition; DrosoNet groups trained on different image regions benefit from extrinsic differentiation, while units within each group benefit from stochastic intrinsic differences.

The outputs of all DrosoNets are then aggregated using a novel voting module which considers multiple top place candidates from each DrosoNet, allowing the system to converge on the most generally agreed upon reference place, mitigating the individual DrosoNets failing to realize a correct match. 

We present a general setup for our proposed system which outperforms other lightweight VPR techniques across several benchmark datasets, while taking less time to retrieve a match. Furthermore, we also compare our results against high-performing but computationally expensive VPR methods to better situate this work in the literature.

The rest of this paper is organized as follows. Section \ref{rel_work} provides an overview of VPR literature with a focus on lightweight methods. Section \ref{method} details our methodology, starting from a short DrosoNet overview, followed by the image partitioning module, training and inference processes, and finishing with the voting aggregation method. Section \ref{exp_setup} explains our experimental setup, providing insight into the benchmark datasets, evaluation metrics and model settings. Results are presented and discussed in Section \ref{results}. We conclude in Section \ref{conclusions} by summarizing our findings, highlighting key system limitations and possible future work.

\section{Related Work}
\label{rel_work}

As the appearance of a place can vary substantially due to a wide variety of environmental and navigation factors, computing an image representation resilient against such changes becomes foundational for autonomous long-term navigation. Nevertheless, the computation, storage, and search of place representations should remain computationally efficient when the target robotic platform cannot afford to carry high-end hardware.

The first image descriptors used for VPR were based on handcrafted methods such as Histogram-of-oriented gradients (HOG) \cite{ref:hog}, which has been successfully used as a global image descriptor for VPR \cite{ref:mcmanus2014scene}. Moreover, when combined with image region-of-interest detectors such as \cite{ref:sift, ref:surf}, HOG acted as a local feature descriptor for VPR.

Machine learning techniques have became increasingly popular in the computer vision community over recent years, and CNN-based methods have crept into VPR applications, achieving high performance when dealing with both appearance changes \cite{zaffar2019levelling} and viewpoint variations \cite{zaffar2019aerialrobotics}. The image descriptors produced by the inner layers of CNNs, even when the model was trained for a different task, are effective in matching place images \cite{ref:cnnimagenet}. When trained specifically for the VPR problem \cite{zhou2017places}, such as HybridNet and AMOSNet \cite{ref:hybridasmosnet}, these CNN-based descriptors achieve even higher VPR performance. With the continuous focus on absolute VPR reliability, these techniques have become increasingly complex. NetVLAD \cite{ref:netvlad} separates the processes of CNN feature extraction and aggregation into two stages. Patch-NetVLAD \cite{patch-netvlad} introduces yet another stage during descriptor matching. While these algorithmic variations and additions do result in increased VPR reliability, the computational cost of such methods prohibits their use with mobile robotics equipped with resource-constrained hardware.
\begin{figure}[!t]
    \centering
    \includegraphics[width=0.9\columnwidth]{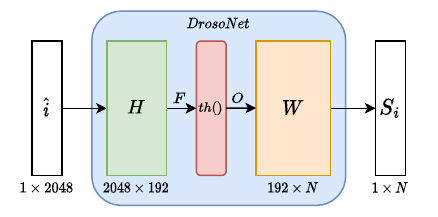}
    \caption{DrosoNet model diagram.}
    \label{fig:droso_diagram}
\end{figure}
Several computationally efficient VPR methods have been proposed to address the shortcomings of CNNs. CoHOG \cite{ref:cohog} was proposed as an efficient and trainless algorithm for VPR. It finds regions-of-interest within an image and computes a HOG descriptor for each found region. CNN adaptations have been proposed to lower their computationally requirements. CALC \cite{ref:calc} is a lightweight CNN-based VPR method which presents lower computational requirements. MobileNets \cite{ref:mobilenets} introduces depth-wise convolutions to lower overall computational requirements. Quantization of neural networks \cite{quant2} into lower bit precisions has also been shown to improve computational profiles. These concepts have been bridged to VPR, with binary neural networks combined with depthwise convolutions \cite{bruno_depthwisebinary} showing great computationally efficiency when paired with specialized hardware.

Efficient bio-inspired VPR methods are designed to mimic the neural activity of small animals, which exhibit incredible navigation capabilities relatively to the size of their brains \cite{ref:honeybee, ref:ants}. RatSLAM \cite{ref:ratslam} takes inspiration from the neural activations of rats to perform navigation. FlyNet \cite{ref:flynetcann} takes inspiration from the brain of the fruit fly \cite{ref:ogflynet} and its odour processing to performs highly efficient VPR. In the authors' previous work, a new algorithm also inspired by the fruit fly was introduced, dubbed DrosoNet \cite{voting_arcanjo}, using multiple of these small models as voting units to perform highly lightweight VPR.

Despite the efforts in developing lightweight VPR techniques, the absolute VPR performance of such methods remains unreliable. In this work, we propose a new approach to a multi-DrosoNet localization system, dubbed RegionDrosoNet, which aims to substantially improve absolute VPR reliability while remaining computationally efficient.

\section{Methodology}
\label{method}

In the interest of self-containment, this section starts by providing a technical background into the DrosoNet model. Following, we detail the proposed image partitioning module, which produces several heterogeneous image regions. The DrosoNet training and inference processes are then described. Finally, the voting module, responsible for aggregating the outputs of all DrosoNets into a final place prediction, is detailed.

\subsection{DrosoNet}

DrosoNet is a compact and fast neural network image classifier where each of the environment's total $N$ places is a different class. We use the same configuration as in \cite{voting_arcanjo}, which can be seen in Fig. \ref{fig:droso_diagram}. An $64\times 32$ grayscale image is first flattened into a one-dimensional vector, denoted as $\hat{i}$, followed by a matrix multiplication with $H$, producing vector $F$. $H$ is a binary, sparse, and randomly initialized matrix, where $10\%$ of each column's elements are initialized to $1$ and the remaining to $0$. $F$ is then binarized by the function $th$, where the top $50\%$ of values are set to $1$ and the bottom $50\%$ are set to $0$, resulting in the binary vector $O$. $W$ is a fully connected layer which learns to map $O$ to one of the $N$ classes, i.e. reference places. The final output vector $s$ stores the score distribution for each reference place, and the DrosoNet's prediction is the index of the largest score in $s$.

While DrosoNet is a fast algorithm, its standalone VPR performance is too unreliable. Moreover, due to the randomness of its $H$ matrix initialization and supervised training, different DrosoNets exhibit high variance in their VPR performance. Combining multiple DrosoNets was hence proposed as an avenue to improve overall VPR performance, relying only the native stochastic behaviour of the models for differentiation \cite{voting_arcanjo}.

This work increases DrosoNet differentiation by training distinct models on different partitions of the original images, producing region specialized DrosoNets. Moreover, by training multiple DrosoNets on each image region, we continue taking advantage of the intrinsic randomness associated with the initialization and training processes.

\subsection{Image Partitioning}

The image partitioning module receives as inputs an image $i$ and grid dimensions $(r, c)$, where $r$ represents the number of rows and $c$ the number of columns, outputting $rc$ image regions. As detailed, DrosoNet operates with grayscale images with a resolution of $64\times 32$, thus the produced regions are converted to grayscale and resized to the correct dimensions.
\begin{figure}[!t]
    \centering
    \includegraphics[width=0.85\columnwidth]{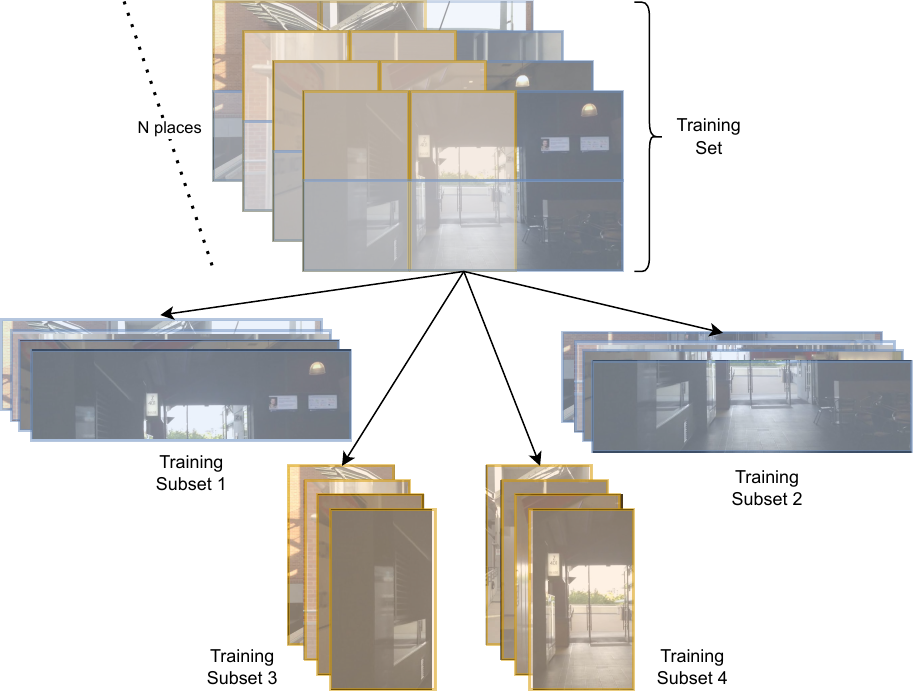}
    \caption{A training subset is produced for each grid position. In this example, the grids $[(2 \times 1), (1 \times 3)]$ are used, with the blue regions highlighting the $2\times1$ grid and the yellow regions the $1\times 3$ grid (the last column was omitted for visibility). The total number of regions is $5$.}
    \label{fig:training}
\end{figure}
\begin{figure*}[thpb]
\vspace*{1ex}
\centering
\includegraphics[width=0.9\textwidth]{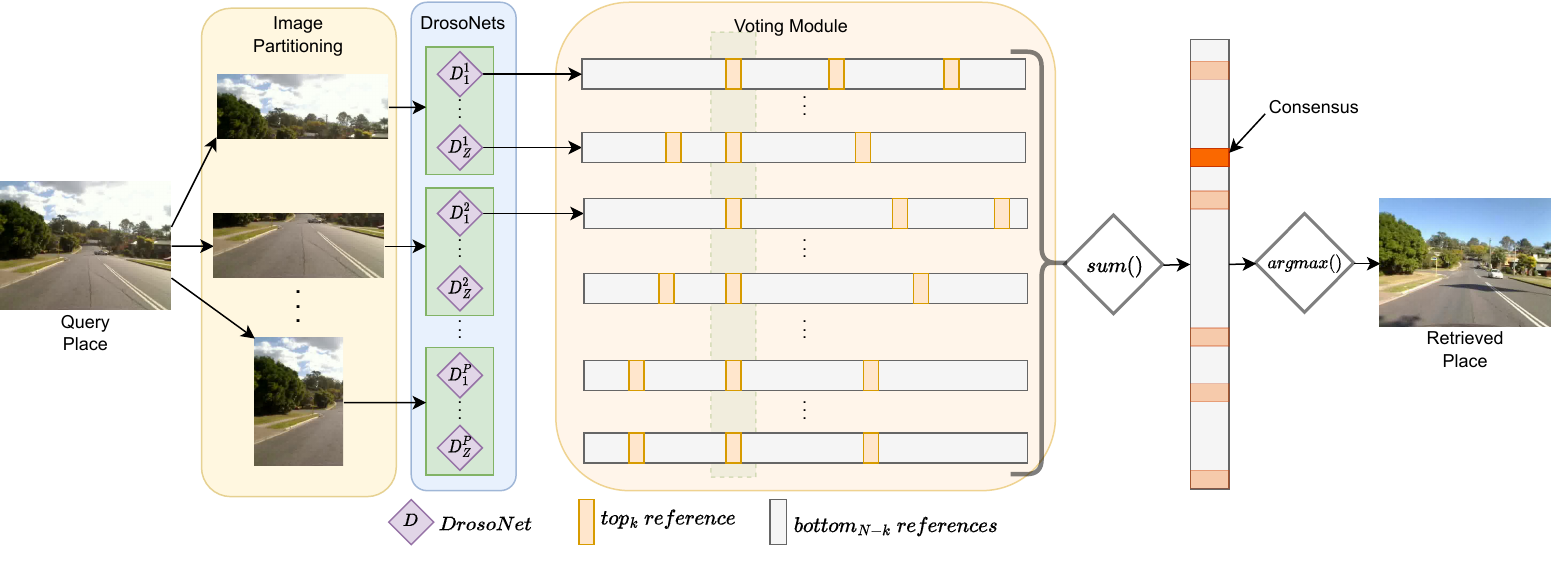}
\caption{The voting module receives all score vectors produced by each DrosoNet, with the largest $K$ values being considered (in this case $K=3$) and all remaining $N-K$ values being discarded.}
\label{votingdiagram}
\end{figure*}
In Section \ref{ablation}, we show how different grid setups can significantly impact the VPR performance of the overall system. Since it is not possible to predict which grid layout is best for the deployment environment without access to ground-truth information, we propose the use of multiple, heterogeneous image regions. In this arrangement, the partitioning process is simply repeated for $G$ different grid settings. The total number of image partitions $P$ can thus be computed as follows:
\begin{gather}
    P = \sum_{g=1}^{G} r_gc_g
    \label{region_calc}
\end{gather}
\noindent where $r_g$ and $c_g$ represent the number of rows and columns associated with grid setup $g$, respectively.

\subsection{Training and Inference}

Each dataset contains $N$ image, one per place, in their training traversal. Before the training process, we construct $P$ training subsets, each corresponding to one of the desired regions (Fig. \ref{fig:training}). Each subset therefore also contains $N$ image partitions.

A group of $Z$ DrosoNets is assigned for each of the $P$ training subsets, with each group being trained only on their respective grid position. The total number of DrosoNets in the system $T$ is therefore given as:
\begin{gather}
    T = PZ
\end{gather}
At inference time, the query image is partitioned following the same $G$ grids, and each DrosoNet is fed the corresponding region of its group, resulting in $T$ score vectors for the query image. All these vectors are aggregated into a final prediction using the proposed voting module.

\subsection{Voting Module}

The voting scheme combines all the output score vectors into a final score vector from which the reference place can be identified. Fig. \ref{votingdiagram} illustrates the matching process for a single query image.

For each of the $T$ score vectors $s$, the voting vector $\hat{s}$ is constructed by setting each of the $N$ elements $\hat{s}_n$ as:
\begin{gather}
  \hat{s}_n =
  \begin{cases}
                                   s_n & \text{if $s_n \geq top_K(s)$ } \\
                                   0 & \text{else} 
  \end{cases}
\end{gather}
\noindent where $top_K(S)$ represents the value of the $K^{th}$ largest score in $s$, with $K$ being an hyperparameter. Fig. \ref{votingdiagram} shows an example of this operation with $K=3$, where only the highest $3$ scores per DrosoNet are considered and the remaining $N-K$ are set to $0$. All the voting vectors are then summed element wise into the final score vector $V$:
\begin{gather}
    v = \sum_{t=1}^{T} \hat{s}^t
\end{gather}
\noindent and the retrieved reference place $m$ is the most voted for index: $m = argmax(v)$.

\section{Experimental Setup}
\label{exp_setup}

This Section details our experimental setup, starting with a presentation of the benchmark datasets, followed by evaluation metrics, comparison VPR methods and implementation settings of our proposed method.
\begin{figure*}[!t]
	\centering
	\begin{subfigure}[b]{0.32\textwidth}
		\centering
		\includegraphics[width=\linewidth]{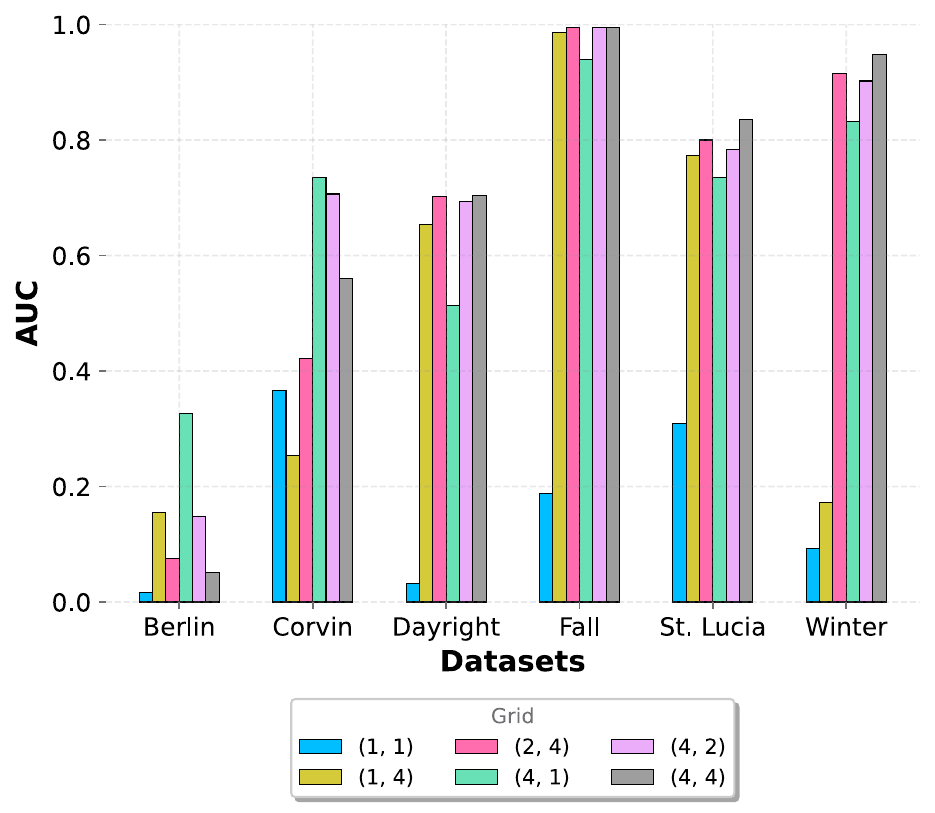}
		\caption{}
		\label{fig:ablation:A}
	\end{subfigure}
	\hfill
	\begin{subfigure}[b]{0.32\textwidth}
		\centering
		\includegraphics[width=\linewidth]{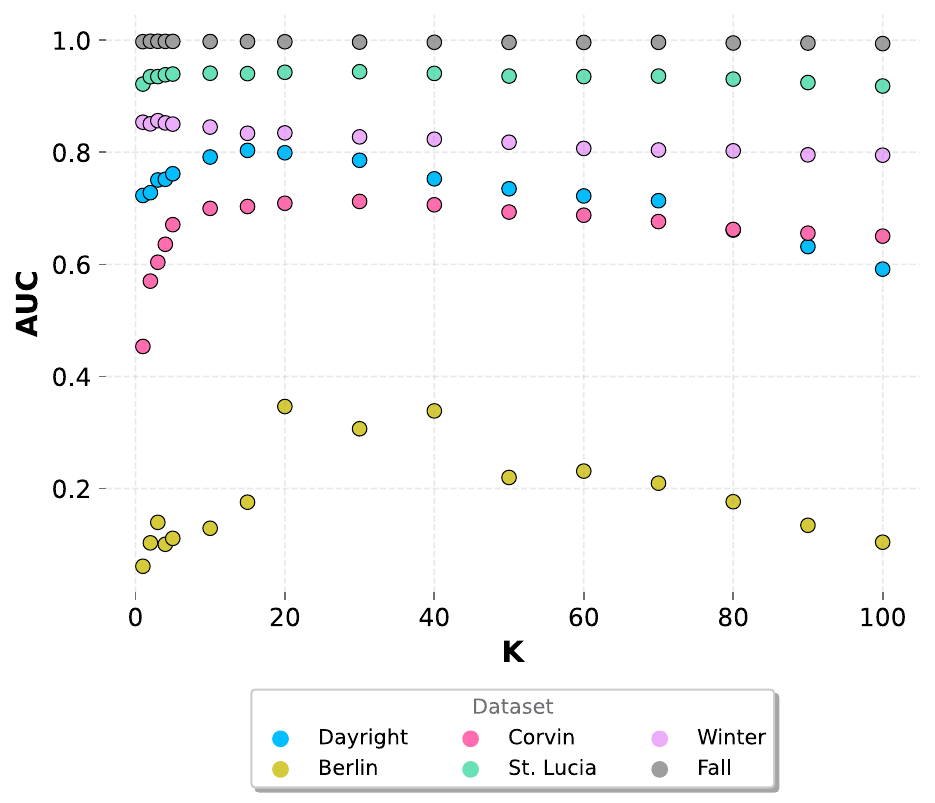}
		\caption{}
		\label{fig:ablation:B}
	\end{subfigure}
	\hfill	
	\begin{subfigure}[b]{0.32\textwidth}
		\centering
		\vspace{2ex}
		\includegraphics[width=\linewidth]{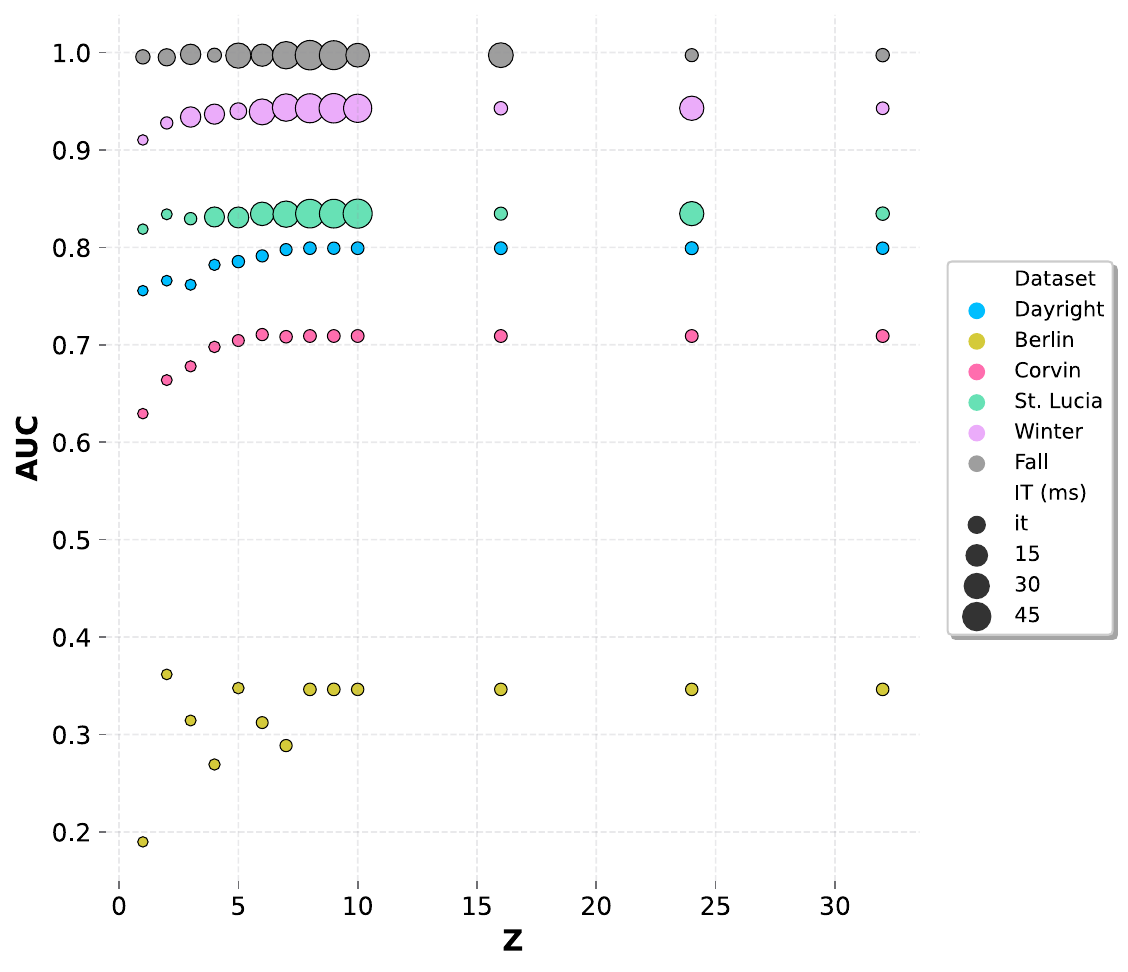}
		\caption{}
		\label{fig:ablation:C}
	\end{subfigure}	
        \caption{AUC impact of the region grid (\ref{fig:ablation:A}), the top $K$ voted places (\ref{fig:ablation:B}) and the number of DrosoNets per region $Z$ (\ref{fig:ablation:C}).}
	\label{fig:ablations}
\end{figure*}
\subsection{Datasets}
\subsubsection{Nordland Fall \& Winter}
The Nordland dataset \cite{ref:nord} consists of four train traversals with varying seasonal weather conditions. We use the Summer traversal as reference for training, testing on the Fall traversal to assess resilience against moderate appearance changes and on the Winter traversal to assess performance with extreme appearance changes. We use $1000$ images per traversal, allowing for a margin for error of $1$ frame around the ground-truth location.

\subsubsection{Gardens Point Day-Right}
The Gardens Point dataset \cite{gardens} consists of three traversals around the Queensland University of Technology. We use the traversal filmed from a left viewpoint during the day as training and the right viewpoint daily traversal as testing, assessing resilience against moderate lateral shifts. The entire $200$ images per traversal are utilized, with an error allowance of $2$ frames.

\subsubsection{St. Lucia}
St. Lucia \cite{stlucia} contains a number of car recorded sequences in St. Lucia, Bribane at different day times. The dataset exhibits moderate appearance changes and dynamic elements. We use the afternoon traversal as reference and the morning sequence as query, with $1150$ images per traversal and an error margin of $2$ frames around the ground-truth location.

\subsubsection{Berlin}
The Berlin dataset \cite{berlin} contains traversals over three locations in Berlin: Halense Strasse, Kudamm and A100. The dataset is characterized by moderate to strong point of view variations and significant dynamic elements such as cars and pedestrians. Due to the small number of frames in each traversal, we combine the three locations into a single dataset, utilizing the traverses halensestrasse-2, kudamm-1 and A100-1 as references and halensestrasse-1, kudamm-2 and A100-2 as queries, resulting in a total of $250$ images. We allow for an error margin of $1$ frame.

\subsubsection{Corvin 30 Degrees}
Corvin \cite{corvin} is a synthetic dataset recorded using flight simulation around the Corvin Castle, focusing on strong viewpoint and scale variations. We use $1000$ images per traversal, with the one filmed at a $0$ degree angle for training and the $30$ degree traversal for testing, allowing for a ground-truth error margin of $20$ frames. Corvin is a challenging dataset and a large error allowance is required to make results for all techniques conclusive \cite{voting_arcanjo}.

\subsection{Evaluations Metrics}
\subsubsection{Area Under The Precision-Recall Curve (AUC)}
AUC is a widely used metric for assessing VPR performance \cite{ref:pr_just2}. In our experiments, we compute Precision-Recall pairs by varying the confidence threshold for which a technique considers a match correct \cite{ref:vpr_bench}. There is usually an inverse relationship between Precision and Recall, and thus the area under the plotted curve is a strong indicative of VPR performance \cite{ref:pr_jus1}. A high AUC value is most useful for applications where retrieving enough possible correct matches is more important than assuring every retrieved match is absolutely correct \cite{vpr_tuto}.

\subsubsection{Extended Precision (EP)}
The Recall at 100\% Precision ($R_{P100}$) metric \cite{ref:cnn_for_vpr2} computes how many correct matches are retrieved before an incorrect one is introduced. It is useful for applications where a single incorrect match would result in catastrophic failure but does not consider the lower performance bound of the technique. EP \cite{EP} combines $R_{P100}$ with the Precision at Minimal Recall, providing a more balanced performance view for such applications.

\subsubsection{Inference Time (IT)}
We measure IT as the time elapsed from the technique receiving a query image to a match being computed. This includes the time required for any runtime image pre-processing, descriptor computation and descriptor matching. We compute IT on the St. Lucia dataset, taking the average of $1100$ inferences. We compute these results on an Intel 12900k processor, running Ubuntu 20.03. The tests are purposely ran without a GPU, as many lower performance robots do not carry an on board dedicated GPU.

\subsection{Comparison VPR Techniques}
We compare RegionDrosoNet to several VPR techniques which claim computational efficiency as one of their main strengths: CALC \cite{ref:calc}, CoHOG \cite{ref:cohog}, and Voting \cite{voting_arcanjo}. Moreover, to better situate our work, we additionally include comparison against the computationally expensive VPR algorithms HybridNet \cite{ref:hybridasmosnet} and Patch-NetVLAd \cite{patch-netvlad}. We use the implementations in \cite{ref:vpr_bench} for CALC, CoHOG and HybridNet, and \cite{vpr_tuto} for Patch-NetVLAD. For Voting, we test both the implementation given in \cite{voting_arcanjo} with $32$ DrosoNets and an additional setup with $82$ to match the same number of DrosoNets as our proposed setup.
\begin{figure*}[!t]
	\centering
	\begin{subfigure}[b]{0.32\textwidth}
		\centering
		\includegraphics[width=\linewidth]{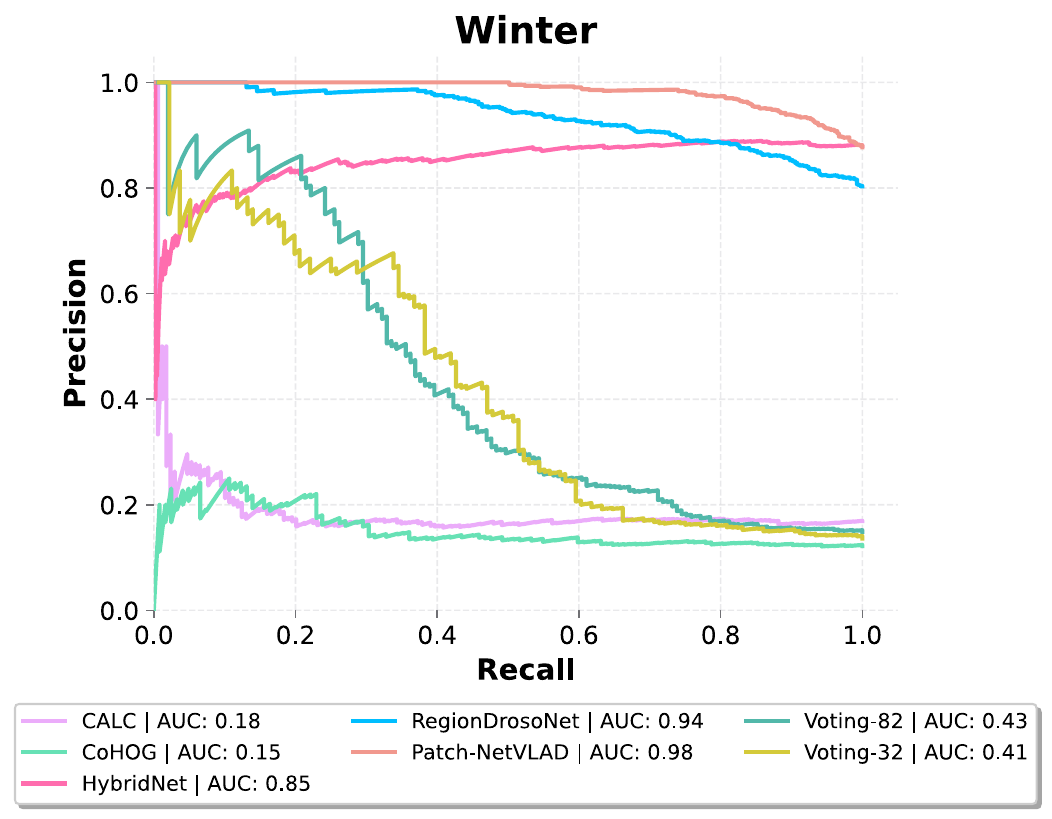}
		\caption{}
		\label{fig:pr_curves:A}
	\end{subfigure}
	\hfill
	\begin{subfigure}[b]{0.32\textwidth}
		\centering
		\includegraphics[width=\linewidth]{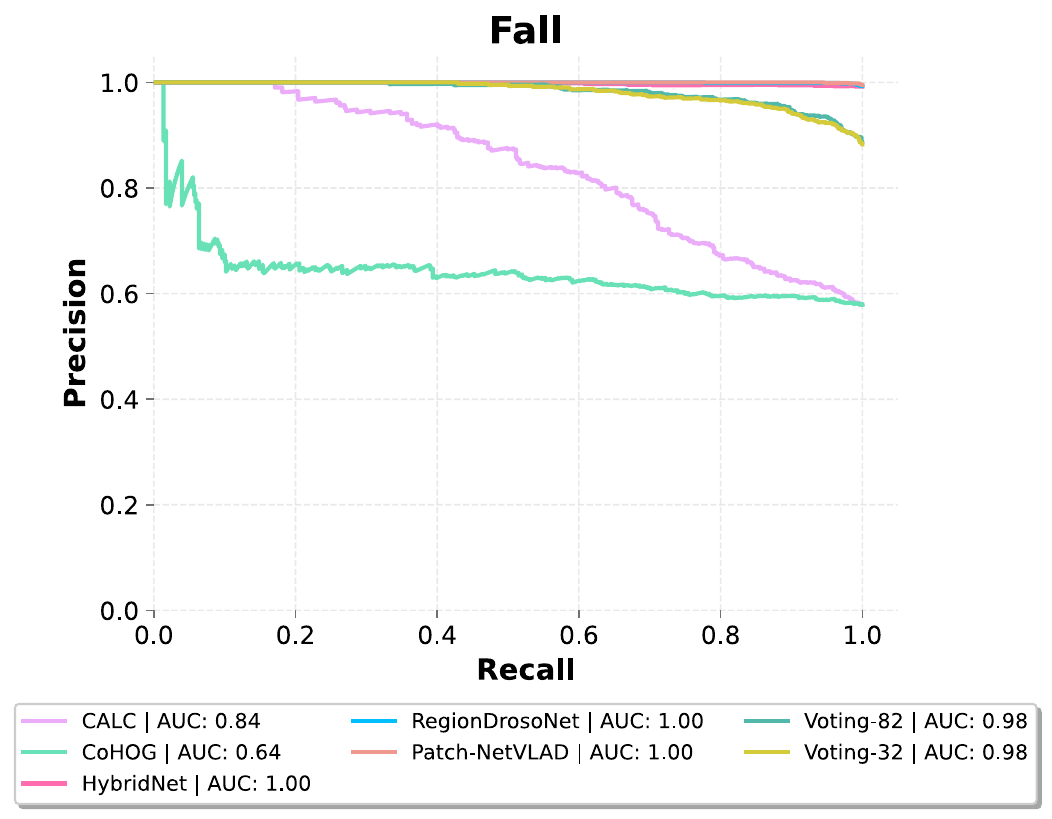}
		\caption{}
		\label{fig:pr_curves:B}
	\end{subfigure}
	\hfill	
	\begin{subfigure}[b]{0.32\textwidth}
		\centering
		\vspace{2ex}
		\includegraphics[width=\linewidth]{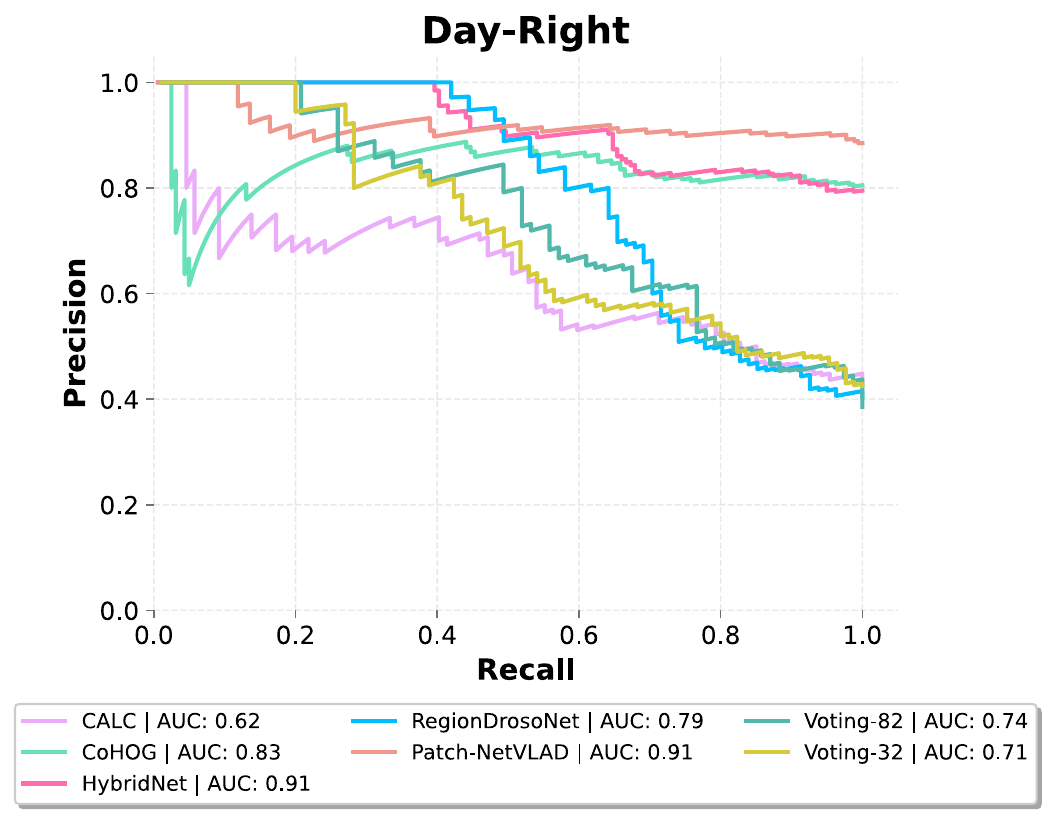}
		\caption{}
		\label{fig:pr_curves:C}
  
	\end{subfigure}	
        \begin{subfigure}[b]{0.32\textwidth}
		\centering
		\includegraphics[width=\linewidth]{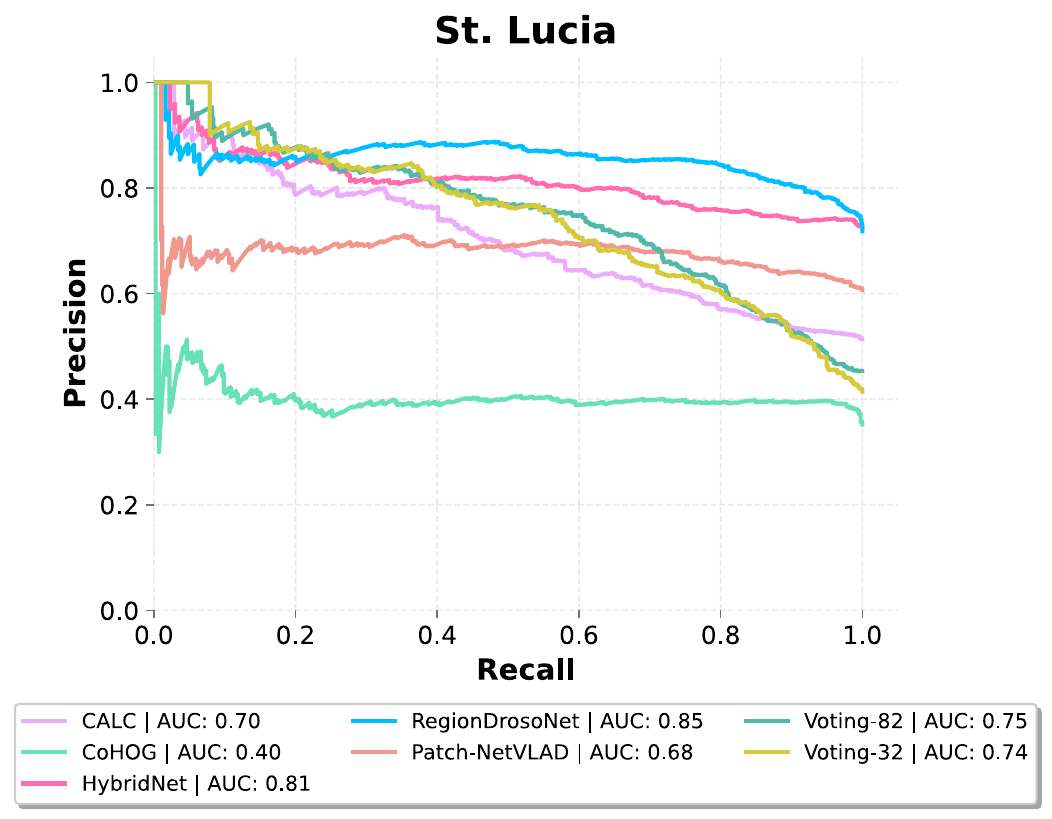}
		\caption{}
		\label{fig:pr_curves:D}
	\end{subfigure}
	\hfill
	\begin{subfigure}[b]{0.32\textwidth}
		\centering
		\includegraphics[width=\linewidth]{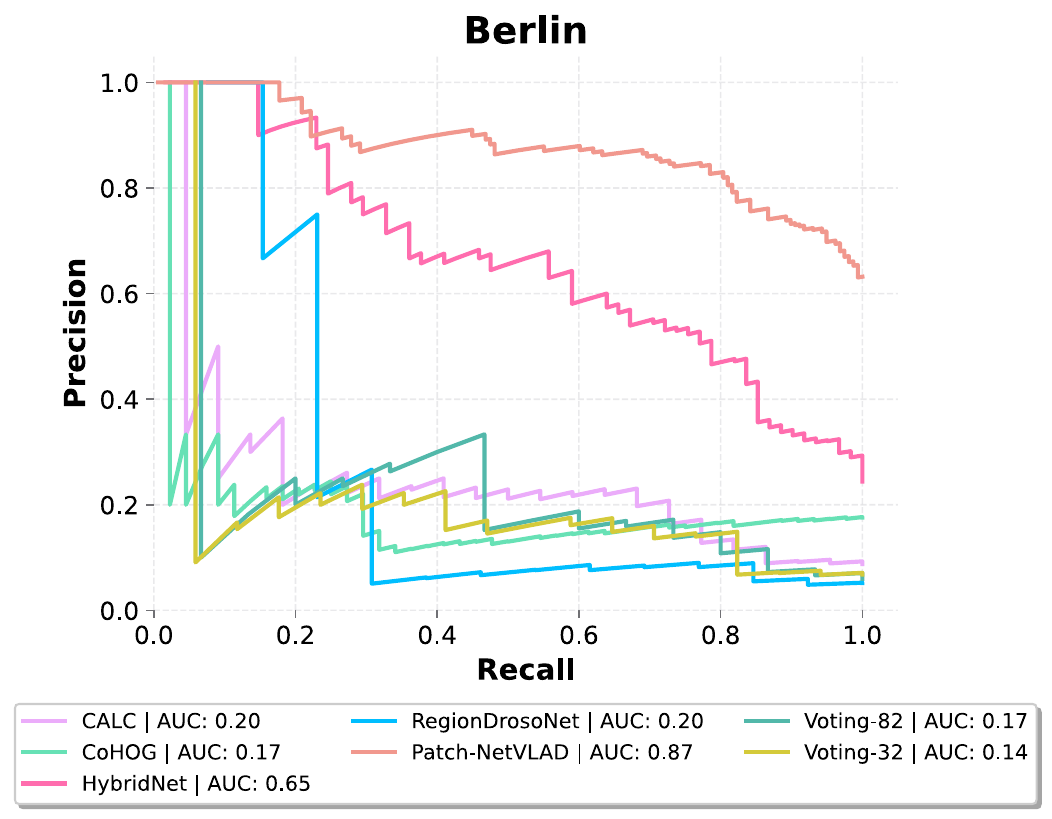}
		\caption{}
		\label{fig:pr_curves:E}
	\end{subfigure}
	\hfill	
	\begin{subfigure}[b]{0.32\textwidth}
		\centering
		\vspace{2ex}
		\includegraphics[width=\linewidth]{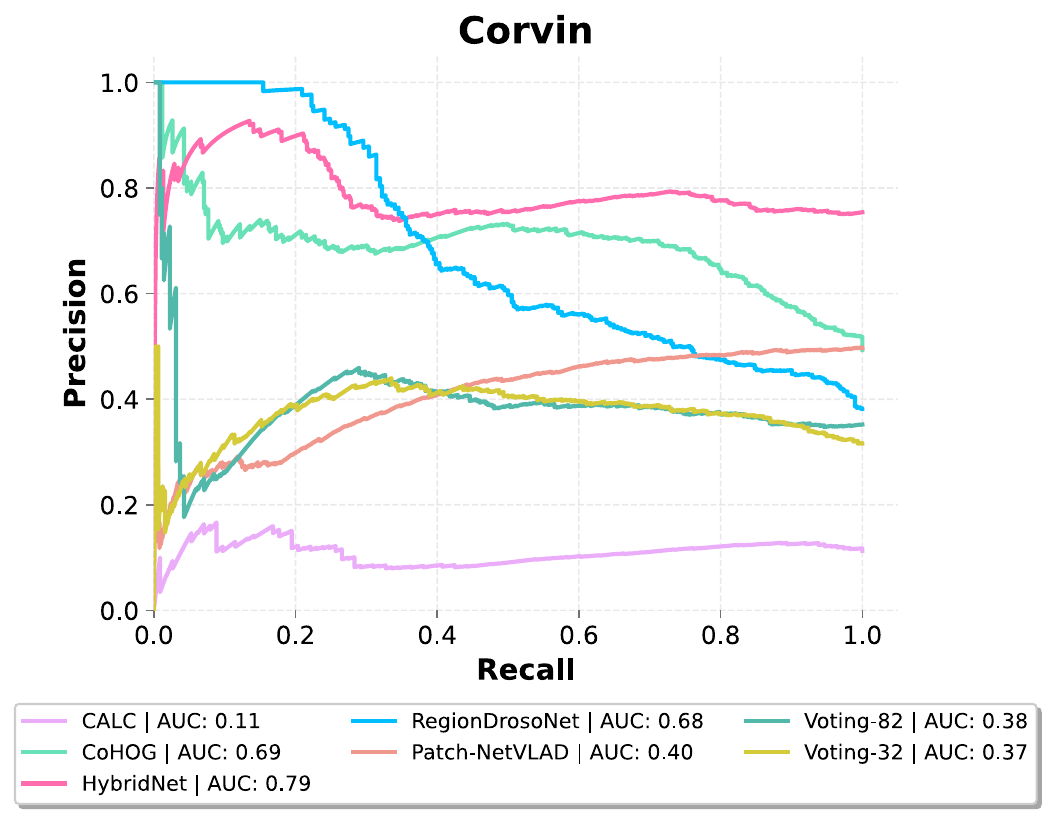}
		\caption{}
		\label{fig:pr_curves:F}
	\end{subfigure}	
	\caption{Precision-recall curves and respective AUC}
	\label{fig:pr_curves}
\end{figure*}
\subsection{Ablation Studies \& Implementation Details}
\label{ablation}

RegionDrosoNet has three main hyperparameters: the grid setups used to construct image regions, the number of DrosoNets per region $Z$, and the number of $top_K$ voted places per DrosoNet. We conduct ablation studies to find optimal settings with the aim of providing a general setup that performs strongly across all datasets, rather than fine-tuning the system for each scenario. The results of these studies can be seen in Fig. \ref{fig:ablations}.

As can be seen in Fig. \ref{fig:ablation:A}, different grid settings significantly impact VPR performance, and the optimal individual grid setting varies from dataset to dataset. As such, we use a combination of all tested partitioning grids:
\begin{gather}
    [(1, 1), (1, 4), (4, 1), (2, 4), (4, 2), (4, 4)]
    \label{grids}
\end{gather}
\noindent resulting in a total of $41$ partitions, following the example scheme in Fig. \ref{fig:training}.

The choice for $K$ also has a substantial impact on VPR performance, as can be seen in Fig. \ref{fig:ablation:B}. We set $K=20$ as it presents the best overall AUC performance across all datasets.

Finally, the number of DrosoNets per region $Z$ has a significant impact on both AUC performance and inference time, observable in Fig. \ref{fig:ablation:C}. We set the system to $Z=2$, as there are heavily diminishing AUC returns with higher $Z$ values, even lowering VPR performance on Corvin and Berlin. With the choice of grids described above, the total number of DrosoNets in the system becomes $82$.

Each DrosoNet is trained for $200$ epochs using the Adam optimizer \cite{kingma2014adam} and with a learning rate of $0.001$.

\section{Results}
\label{results}
This section presents and discusses our results, firstly with a comparison of RegionDrosoNet versus other computationally efficient VPR techniques, followed by a comparison against expensive methods and finalizing with a per-region performance analysis.

\subsection{VPR Performance VS Lightweight Methods}

In Fig. \ref{fig:pr_curves} we observe the VPR performance in terms of AUC for all tested techniques. RegionDrosoNet outperforms every other lightweight algorithm on all appearance-based datasets (Winter, Fall and St. Lucia). The performance advantage on the Winter dataset over other efficient methods is the most notable, with RegionDrosoNet more than doubling the AUC of the second best efficient technique (Voting-82).
Viewpoint performance on the Corvin dataset is also commendable, with RegionDrosoNet achieving the highest EP result and matching CoHOG in AUC. While all lightweight techniques perform poorly on the Berlin dataset, our method achieves the highest EP amongst them and ties with CALC for the highest AUC. The VPR performance of Voting-32 and Voting-82 is functionally indistinguishable, showing that simply increasing the number of DrosoNets does not contribute significantly to place matching. Conversely, the use of 82 units in the proposed pipeline provides significant improvements in VPR, as demonstrated by the performance gap between RegionDrosoNet and Voting-82.
\begin{figure}[!t]
    \centering
    \includegraphics[width=0.9\columnwidth]{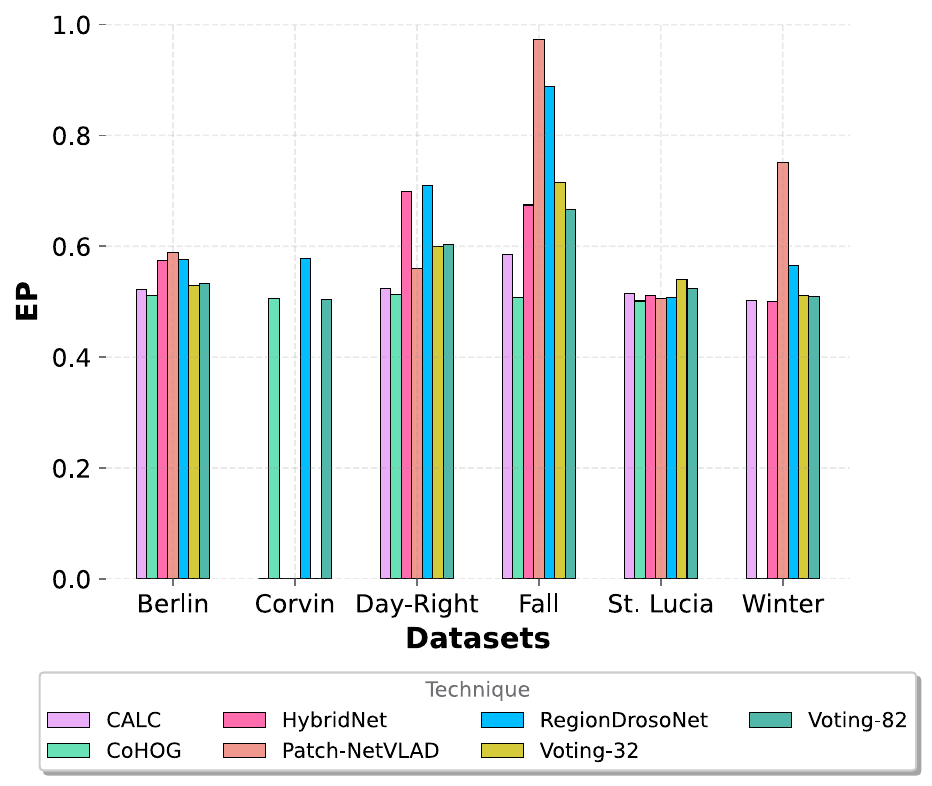}
    \caption{Extended precision (EP) comparison.}
    \label{fig:eps}
\end{figure}
Table \ref{memory&time} shows the inference times at runtime for every tested technique. RegionDrosoNet is the third-fastest method, second only to Voting-32 and Voting-82, the latter due to the extra image pre-processing required by RegionDrosoNet. Nevertheless, it achieves substantially higher VPR reliability on both viewpoint and appearance-based visual challenges while remaining $18$ times faster than CALC and over two orders of magnitude faster than CoHOG.

Despite these efficiency advantages, it is worth noting that different methods can offer various benefits over each other. CoHOG, while requiring the reference traversal images for the reference map computation, is a trainless technique. CALC, while trained and also requiring the reference place images for the descriptor database, does not require environment specific training. RegionDrosoNet, while achieving better VPR performance and efficiency, does require environment specific training due to its dependency on DrosoNet. The choice of a VPR technique is highly application dependant and all factors such as data availability, hardware, deployment environment and risk of failure should be taken into account.

\subsection{VPR Performance VS Expensive Methods}
As can be seen in Table \ref{memory&time}, HybridNet and Patch-NetVLAD are significantly slower than the lightweight methods. 

Despite its substantially lower computational requirements, RegionDrosoNet is able to compete with these expensive methods, even outperforming them on some datasets. On the Corvin dataset, RegionDrosoNet achieves higher EP. In the challenging Winter dataset, it outperforms HybridNet in both EP and AUC. The highest performance drop from RegionDrosoNet is in Berlin, where it loses substantially in both AUC and EP to the costly techniques.
\newcolumntype{M}[1]{>{\centering\arraybackslash}m{#1}}
\begin{table}[t]
  \centering
  \caption{Inference Time (IT) \& Frames Per Second (FPS)}
  \label{memory&time}%
    \begin{tabular}{M{2cm}M{1.4cm}M{1.4cm}}
    \toprule
Model     & IT (ms) & FPS   \\
\midrule
CoHOG     & 671              & 1.49      \\
CALC      & 166               & 6.02       \\
Voting-32    & 3                & 333.33    \\
Voting-82 & 8 & 125.00 \\
HybridNet & 3318              & 0.30       \\  
Patch-NetVLAD   & 2892          & 0.35    \\
RegionDrosoNet & 9 & 111.11 \\
\bottomrule
\end{tabular}
\end{table}%

\subsection{Per-Region Insights}

\begin{figure}[!t]
	\centering
	\begin{subfigure}[b]{0.40\textwidth}
		\centering
		\includegraphics[width=\linewidth]{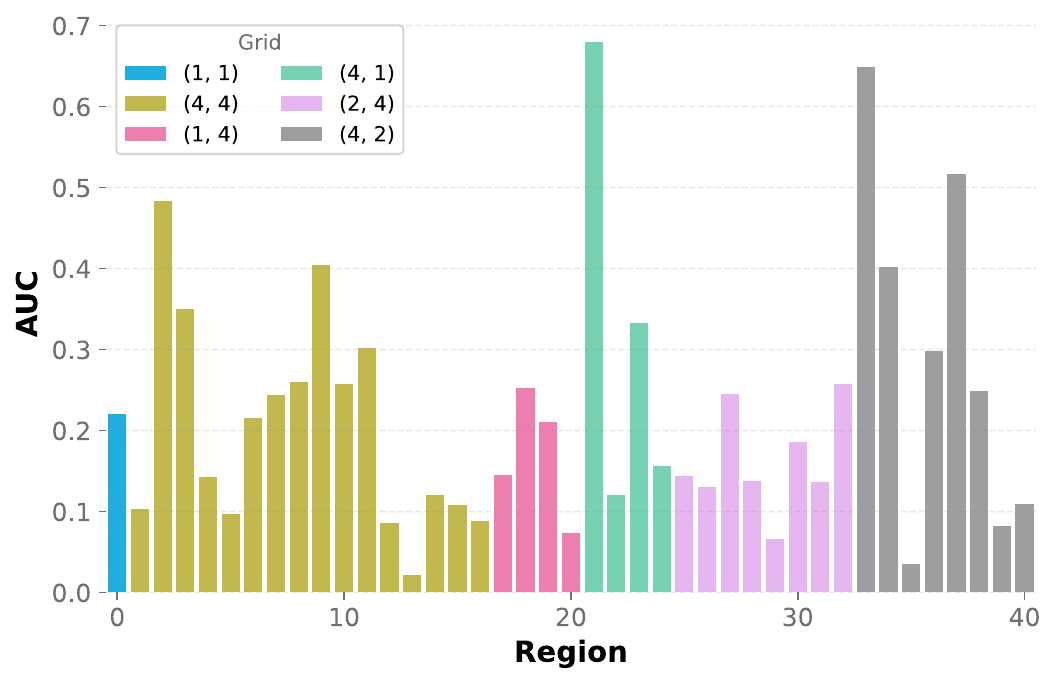}
		\caption{Corvin}
		\label{fig:aucpr:A}
	\end{subfigure}
	\hfill
	\begin{subfigure}[b]{0.40\textwidth}
		\centering
		\includegraphics[width=\linewidth]{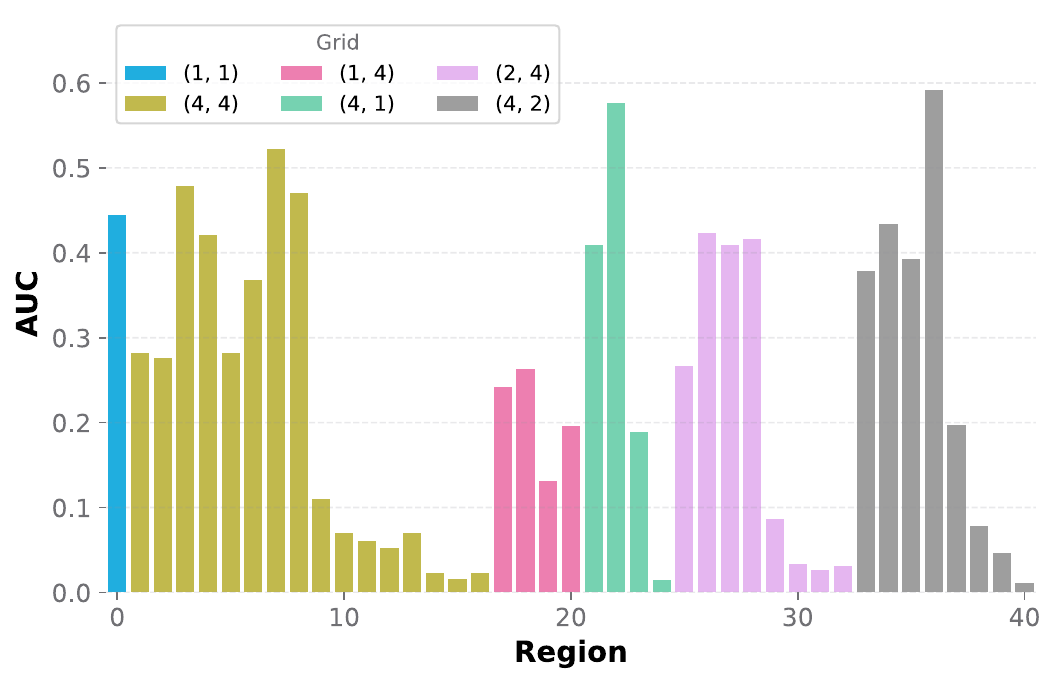}
		\caption{St. Lucia}
		\label{fig:aucpr:B}
	\end{subfigure}

        \caption{AUC per region, with colour highlighting the associated grid dimensions. Within each grid, regions are placed from left-to-right, top-to-bottom. E.g., for grid $(4, 4)$, its first bar represents $row 1, \, column 1$, the second bar $row 1, \, column 2$, the fifth bar $row2, \, column 1$, etc.}
	\label{fig:aucperpart}
\end{figure}

In Fig. \ref{fig:aucperpart} we show RegionDrosoNet's AUC per region on the Corvin (\ref{fig:aucpr:A}) and St. Lucia (\ref{fig:ablation:B}) datasets. As per Eq. \ref{region_calc} and Eq. \ref{grids}, our setup has a total of $41$ regions, each represented by a bar, where the colour code shows the corresponding grid arrangement from which it originated from. It is clear that some regions perform substantially better than others, and region performance is dataset dependant. Furthermore, the region corresponding to the whole query image (region $0$, in blue) is not the best performing one.

\begin{figure}[t]
    \centering
    \includegraphics[width=0.9\columnwidth]{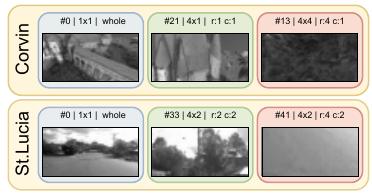}
    \caption{Query regions: whole image in blue, best performing region in green, and worse performing region in red.}
    \label{fig:corvingoodvbad}
\end{figure}

Looking at Fig. \ref{fig:corvingoodvbad}, we find visual insights for the large performance discrepancy. On Corvin, region $13$ does not have enough visual detail for DrosoNet to specialize on, while $21$ contains strong features. Region $13$ also performs better than the whole query image, as the former has less non-detailed visual zones and less compression resulting from the image scaling pre-processing. Finally, St. Lucia follows the same pattern with its respective best and worst performing regions.

\section{Conclusions and Future Work}
\label{conclusions}

In this work, we propose RegionDrosoNet: a novel multi-DrosoNet localization system which significantly improves upon the VPR performance of current lightweight methods while remaining computational efficient. The approach relies on increasing the differentiation of different DrosoNets by training specialized groups on several image partitions. Moreover, the introduce a novel voting method which considers multiple top place candidates from each DrosoNet, allowing a correct consensus to be reached even if individual DrosoNets place an incorrect highest scoring match.

DrosoNet is a neural network classifier which requires training on the reference set of the target environment. While training time is low compared to expensive models, it remains a limitation of this work. Future research could focus on adapting DrosoNet into a descriptor-based method which does not require environment specific training.





\bibliographystyle{IEEEtran}
\typeout{}
\bibliography{ref}

\end{document}